\documentclass[]{article} 
\usepackage{colacl}

\begin{document}

\title{A Compact Architecture for Dialogue Management Based on Scripts and Meta-Outputs }
\author{Manny Rayner, Beth Ann Hockey, Frankie James\\ 
Research Institute for Advanced Computer Science\\ 
Mail Stop 19-39, NASA Ames Research Center\\
Moffett Field, CA 94035-1000\\
\{manny,bahockey,fjames\}@riacs.edu}

\maketitle

\begin{abstract}
We describe an architecture for spoken dialogue interfaces to
semi-autonomous systems that transforms speech signals through
successive representations of linguistic, dialogue, and domain
knowledge.  Each step produces an output, and a meta-output describing
the transformation, with an executable program in a simple scripting
language as the final result. The output/meta-output distinction
permits perspicuous treatment of diverse tasks such as resolving
pronouns, correcting user misconceptions, and optimizing scripts.
\end{abstract}

\section{Introduction\footnote{This paper also appears in the proceedings of the Sixth International Conference on Applied Natural Language Processing, Seattle, WA, April 2000.}}

The basic task we consider in this paper is that of using spoken
language to give commands to a semi-autonomous robot or other similar
system. As evidence of the importance of this task in the NLP
community note that the early, influential system SHRDLU
\cite{Winograd73} was intended to address just this type of problem.
More recent work on spoken language interfaces to semi-autonomous
robots include SRI's Flakey robot \cite{Flakey93} and NCARAI's
InterBOT project \cite{Perzanowski98,Perzanowski99}.  A number of
other systems have addressed part of the task. CommandTalk
\cite{MooreEA97}, Circuit Fix-It Shop \cite{Smith97} and TRAINS-96
\cite{TraumAllen94,TraumAndersen99} are spoken language systems but
they interface to simulation or help facilities rather than
semi-autonomous agents.  Jack's MOOse Lodge \cite{BadlerEA99} takes
text rather than speech as natural language input and the avatars being
controlled are not semi-autonomous.  Other researchers have considered
particular aspects of the problem such as accounting for various
aspects of actions \cite{Weber95,Pym95}.  In most of this and other
related work the treatment is some variant of the following. If there
is a speech interface, the input speech signal is converted into
text. Text either from the recognizer or directly input by the user is
then converted into some kind of logical formula, which abstractly
represents the user's intended command; this formula is then fed into
a command interpreter, which executes the command.

We do not think the standard treatment outlined above is in essence
incorrect, but we do believe that, as it stands, it is in need of some
modification. This paper will in particular make three points. First,
we suggest that the output representation should not be regarded as a
logical expression, but rather as a program in some kind of scripting
language. Second, we argue that it is not merely the case that the
process of converting the input signal to the final representation can
sometimes go wrong; rather, this is the normal course of events, and
the interpretation process should be organized with that assumption in
mind. Third, we claim, perhaps surprisingly, that the first and second
points are related. These claims are elaborated in
Section~\ref{Section:Theoretical}.

The remainder of the paper describes an architecture which addresses
the issues outlined above, and which has been used to implement a
prototype speech interface to a simulated semi-autonomous robot
intended for deployment on the International Space
Station. Sections~\ref{Section:Implementation}
and~\ref{Section:CompactArchitecture} present an overview of the
implemented interface, focussing on representational issues relevant
to dialogue management. Illustrative examples of interactions with the
system are provided in Section~\ref{sec-examples}.
Section~\ref{Section:Conclusions} concludes.

\section{Theoretical Ideas}
\label{Section:Theoretical}

\subsection{Scripts vs Logical Forms}

Let's first look in a little more detail at the question of what the
output representation of a spoken language interface to a semi-autonomous robot/agent should be. In practice, there seem to be two
main choices: atheoretical representations, or some kind of logic.

Logic is indeed an excellent way to think about representing
static relationships like database queries, but it is much less clear
that it is a good way to represent {\em commands}. In real life, when
people wish to give a command to a computer, they usually do so via
its operating system; a complex command is an expression in a
scripting language like CSHELL, Perl, or VBScript. These languages
are related to logical formalisms, but cannot be mapped onto them in a
simple way. Here are some of the obvious differences:

\begin{itemize}

\item A scripting language is essentially imperative, rather than relational.

\item The notion of temporal sequence is fundamental to the
language. ``Do $P$ and then $Q$'' is not the same as ``Make the goals
$P$ and $Q$ true''; it is explicitly stated that $P$ is to be done
first. Similarly, ``For each $X$ in the list $(A\ B\ C)$, do $P(X)$''
is not the same as ``For all $X$, make $P(X)$ true''; once again, the
scripting language defines an order, but not the logical
language\footnote{In cases like these, the theorem prover or logic
programming interpreter used to evaluate the logical formula typically
assigns a conventional order to the conjuncts; note however that this
is part of the {\em procedural} semantics of the theorem
prover/interpreter, and does not follow from the declarative semantics
of the logical formalism.}.

\item Scripting languages assume that commands do not always
succeed. For example, UNIX-based scripting languages like CSHELL
provide each script with the three predefined streams {\tt stdin},
{\tt stdout} and {\tt stderr}. Input is read from {\tt stdin}
and written to {\tt stdout}; error messages, warnings and other
comments are sent to {\tt stderr}.  

\end{itemize}

We do not think that these properties of scripting language are
accidental. They have evolved as the result of strong selectional
pressure from real users with real-world tasks that need to be carried
out, and represent a competitive way to meet said users' needs. We
consequently think it is worth taking seriously the idea that a target
representation produced by a spoken language interface should share
many of these properties.

\subsection{Fallible Interpretation: Outputs and Meta-outputs}

We now move on to the question of modelling the interpretation
process, that is to say the process that converts the input (speech)
signal to the output (executable) representation. As already
indicated, we think it is important to realize that interpretation is
a process which, like any other process, may succeed more or less well
in achieving its intended goals.  Users may express themselves
unclearly or incompletely, or the system may more or less seriously
fail to understand exactly what they mean. A good interpretation
architecture will keep these considerations in mind.

Taking our lead from the description of scripting languages sketched
above, we adapt the notion of the ``error stream'' to the interpretation
process. In the course of interpreting an utterance, the system 
translates it into successively ``deeper'' levels of representation.
Each translation step has not only an input (the representation consumed)
and an output (the representation produced), but also something we
will refer to as a ``meta-output'': this provides information about
how the translation was performed. 

At a high level of abstraction, our architecture will be as follows.
Interpretation proceeds as a series of non-deterministic translation
steps, each producing a set of possible outputs and associated
meta-outputs. The final translation step produces an executable
script. The interface attempts to simulate execution of each possible
script produced, in order to determine what would happen if that
script were selected; simulated execution can itself produce further
meta-outputs. Finally, the system uses the meta-output information
to decide what to do with the various possible interpretations it
has produced. Possible actions include selection and execution of
an output script, paraphrasing meta-output information back to the user,
or some combination of the two.

In the following section, we present a more detailed description
showing how the output/meta-output distinction works in a
practical system.

\section{A Prototype Implementation}
\label{Section:Implementation}

The ideas sketched out above have been realized as a prototype spoken
language dialogue interface to a simulated version of the Personal Satellite
Assistant (PSA;~\cite{PSA}). This section gives an overview of the
implementation; in the following section, we focus on the specific
aspects of dialogue management which are facilitated by the
output/meta-output architecture.

\subsection{Levels of Representation}

The real PSA is a miniature robot currently being developed at NASA
Ames Research Center, which is intended for deployment on the Space
Shuttle and/or International Space Station. It will be capable of free
navigation in an indoor micro-gravity environment, and will provide
mobile sensory capacity as a backup to a network of fixed sensors. The
PSA will primarily be controlled by voice commands through a hand-held
or head-mounted microphone, with speech and language processing being
handled by an offboard processor.  Since the speech processing units
are not in fact physically connected to the PSA we envisage that they
could also be used to control or monitor other environmental
functions. In particular, our simulation allows voice access to the
current and past values of the fixed sensor readings.

The initial PSA speech interface demo consists of a simple simulation
of the Shuttle. State parameters include the PSA's current position,
some environmental variables such as local temperature, pressure and
carbon dioxide levels, and the status of the Shuttle's doors
(open/closed). A visual display gives direct feedback on some of these
parameters.

The speech and language processing architecture is based on that of
the SRI CommandTalk system~\cite{MooreEA97,StentEA99}.  The
system comprises a suite of about 20 agents, connected together using
the SRI Open Agent Architecture (OAA;~\cite{MartinEA98}).  Speech
recognition is performed using a version of the Nuance
recognizer~\cite{Nuance}. Initial language processing is carried out
using the SRI Gemini system~\cite{DowdingEA93}, using a domain-independent
unification grammar and a domain-specific lexicon. The language
processing grammar is compiled into a recognition grammar using the
methods of~\cite{MooreEA97}; the net result is that only
grammatically well-formed utterances can be recognized. Output from
the initial language-processing step is represented in a version of
Quasi Logical Form~\cite{VanEijckMoore92}, and passed in that form to the
dialogue manager. We refer to these as {\em linguistic level}
representations.

The aspects of the system which are of primary interest here concern
the dialogue manager (DM) and related modules. Once a linguistic level
representation has been produced, the following processing steps
occur:

\begin{itemize}

\item The linguistic level representation is converted into a
{\em discourse level representation}. This primarily involves
regularizing differences in surface form: so, for example, ``measure the
pressure'' and ``what is the pressure?'' have different representations
at the linguistic level, but the same representation at the discourse
level. 

\item If necessary, the system attempts to resolve instances
of ellipsis and anaphoric reference. For example, if the previous
command was ``measure temperature at flight deck'', then the new command
``lower deck'' will be resolved to an expression meaning ``measure
temperature at lower deck''. Similarly, if the previous command was
``move to the crew hatch'', then the command ``open it'' will be resolved
to ``open the crew hatch''. We call the output of this step a
{\em resolved discourse level representation}. 

\item The resolved
discourse level representation is converted into an executable script
in a language essentially equivalent to a subset of CSHELL. This
involves two sub-steps. First, quantified variables are given
{\em scope}: for example, ``go to the flight deck and lower deck and
measure pressure'' becomes something approximately equivalent to the
script
\begin{verbatim}
foreach x (flight_deck lower_deck)
  go_to $x
  measure pressure
end
\end{verbatim}
The point to note here is that the {\tt foreach} has scope over both
the {\tt go\_to} and the {\tt measure} actions; an alternate (incorrect)
scoping would be
\begin{verbatim}
foreach x (flight_deck lower_deck)
  go_to $x
end
measure pressure
\end{verbatim}
The second sub-step is to attempt to optimize the plan. In the
current example, this can be done by reordering the list
{\tt (flight\_deck lower\_deck)}. For instance, if the PSA is already at
the lower deck, reversing the list will mean that the robot only
makes one trip, instead of two.

\item The final step in the interpretation process is {\em plan
evaluation}: the system tries to work out what will happen if it
actually executes the plan. (The relationship between plan evaluation
and plan execution is described in more detail in
Section~\ref{Section:PlanEvaluation}).  Among other things, this gives
the dialogue manager the possibility of comparing different
interpretations of the original command, and picking the one which is
most efficient.

\end{itemize}

\subsection{How Meta-outputs Participate in the Translation}
\label{Section:MetaOutputsTranslation}

The above sketch shows how context-dependent interpretation is
arranged as a series of non-deterministic translation steps; in each
case, we have described the input and the output for the step in
question. We now go back to the concerns of
Section~\ref{Section:Theoretical}. First, note that each translation
step is in general fallible. We give several examples:

\begin{itemize}

\item One of the most obvious cases arises when the user simply issues
an invalid command, such as requesting the PSA to open a door $D$ which is
already open. Here, one of the meta-outputs issued by the plan evaluation step 
will be the term
\begin{verbatim}
presupposition_failure(already_open(D));
\end{verbatim}
the DM can decide to paraphrase this back to the user as a surface
string of the form ``$D$ is already open''. Note that plan
evaluation does not involve actually executing the final script, which
can be important. For instance, if the command is ``go to the crew
hatch and open it'' and the crew hatch is already open, the interface
has the option of informing the user that there is a problem without
first carrying out the ``go to'' action. 

\item The resolution step can give rise to similar kinds of meta-output.
For example, a command may include a referring expression that has no
denotation, or an ambiguous denotation; for example, the user might
say ``both decks'', presumably being unaware that there are in fact
three of them. This time, the meta-output produced is
\begin{verbatim}
presupposition_failure(
    incorrect_size_of_set(2,3))
\end{verbatim}
representing the user's incorrect belief about the number of decks.
The DM then has the possibility of informing the user of this misconception
by realizing the meta-output term as the surface string 
``in fact there are three of them''.

Ambiguous denotation occurs when a description is under-specified. For
instance, the user might say ``the deck'' in a situation where there is no
clearly salient deck, either in the discourse situation or in the simulated
world: here, the meta-output will be
\begin{verbatim}
presupposition_failure(
    underspecified_definite(deck))
\end{verbatim}
which can be realized as the clarification question ``which deck do you mean?''

\item A slightly more complex
case involves plan costs. During plan evaluation, the system simulates
execution of the output script while keeping track of execution
cost. (Currently, the cost is just an estimate of the time required to
execute the script). Execution costs are treated as
meta-outputs of the form
\begin{verbatim}
cost(C)
\end{verbatim}
and passed back through the interpreter so that the plan
optimization step can make use of them. 

\item Finally, we consider what happens when the system receives incorrect
input from the speech recognizer. Although the recognizer's language model is
constrained so that it can only produce grammatical utterances, it can still
misrecognize one grammatical string as another one.
Many of these cases fall into one of a small number of syntactic
patterns, which function as fairly reliable indicators of bad recognition. A
typical example is conjunction involving a pronoun: if the system hears
``it and flight deck'', this is most likely a misrecognition of something like
``go to flight deck''. 

During the processing phase which translates linguistic level representations
into discourse level representations, the system attempts to match
each misrecognition pattern against the input linguistic form, and if successful 
produces a meta-output of the form
\begin{verbatim}
presupposition_failure(
    dubious_lf(<Type>))
\end{verbatim}
These meta-outputs are passed down to the DM, which in the absence of
sufficiently compelling contrary evidence will normally issue a
response of the form ``I'm sorry, I think I misheard you''.

\end{itemize}

\section{A Compact Architecture for Dialogue Management Based on Scripts
and Meta-Outputs}
\label{Section:CompactArchitecture}

None of the individual functionalities outlined above are particularly
novel in themselves. What we find new and interesting is the fact that
they can all be expressed in a uniform way in terms of the
script output/meta-output architecture. This section presents three
examples illustrating how the architecture can be used to simplify
the overall organization of the system.

\subsection{Integration of plan evaluation, plan execution and dialogue management}
\label{Section:PlanEvaluation}

Recall that the DM simulates evaluation of the plan before running it,
in order to obtain relevant meta-information. At plan execution time,
plan actions result in changes to the world; at plan evaluation time,
they result in {\em simulated} changes to the world and/or produce
meta-outputs.

Conceptualizing plans as scripts rather than logical formulas permits
an elegant treatment of the execution/evaluation dichotomy. There is
one script interpreter, which functions both as a script executive and
a script evaluator, and one set of rules which defines the procedural
semantics of script actions. Rules are parameterized by execution type
which is either ``execute'' or ``evaluate''.  In ``evaluate'' mode,
primitive actions modify a state vector which is threaded through the
interpreter; in ``execute'' mode, they result in commands being sent to
(real or simulated) effector agents. Conversely, ``meta-information''
actions, such as presupposition failures, result in output being sent
to the meta-output stream in ``evaluate'' mode, and in a null action in
``execute'' mode. The upshot is that a simple semantics can be assigned to
rules like the following one, which defines the action of attempting
to open a door which may already be open:
\begin{verbatim}
procedure(
 open_door(D),
 if_then_else(status(D, open_closed, open),
   presupposition_failure(already_open(D)),
   change_status(D, open_closed, open)))
\end{verbatim}

\subsection{Using meta-outputs to choose between interpretations}

As described in the preceding section, the resolution step is in
general non-deterministic and gives rise to meta-outputs which
describe the type of resolution carried out. For example, consider a
command involving a definite description, like ``open the
door''. Depending on the preceding context, resolution will produce a
number of possible interpretations; ``the door'' may be resolved to
one or more contextually available doors, or the expression may be 
left unresolved. In each case, the type of resolution used appears
as a meta-output, and is available to the dialogue manager
when it decides which interpretation is most felicitous. By default,
the DM's strategy is to attempt to supply antecedents for referring
expressions, preferring the most recently occurring sortally appropriate
candidate. In some cases, however, it is desirable to allow the
default strategy to be overridden: for instance, it may result in 
a script which produces a presupposition failure during plan evaluation.
Treating resolution choices and plan evaluation problems as similar
types of objects makes it easy to implement this kind of idea.

\subsection{Using meta-outputs to choose between dialogue management moves}

Perhaps the key advantage of our architecture is that collecting
together several types of information as a bag of meta-outputs
simplifies the top-level structure of the dialogue manager. In our
application, the critical choice of dialogue move comes after the
dialogue manager has selected the most plausible interpretation.  It
now has to make two choices. First, it must decide whether or not to
paraphrase any of the meta-outputs back to the user; for example, if
resolution was unable to fill some argument position or find an
antecedent for a pronoun, it may be appropriate to paraphrase the
corresponding meta-output as a question, e.g. ``where do you mean?'', or
``what do you mean by `it'?''. Having all the meta-outputs
available together means that the DM is able to plan a coherent
response: so if there are several meta-outputs which could potentially
be worth paraphrasing, it typically realizes only the most important one.

Second, if interpretation was able to produce a well-formed plan, the
DM currently has the three options of executing it, paraphrasing it
back to the user as a confirmation question, or doing nothing. 
Once again, this decision often requires global information about 
what has happened during the interpretation process. For example,
knowing that plan evaluation showed that the plan would take significant
time to execute makes it more plausible that the user would prefer 
to receive a confirmation. 

\section{Examples}
\label{sec-examples}

This section presents a commented sample dialogue with the implemented
PSA interface demo chosen to illustrate the points discussed in the previous sections.  We focus in particular on the role that meta-outputs play in interpretation. Text preceded by USER represents spoken
utterances from the user. Text preceded by PSA represents
synthesized spoken responses from the PSA. Text inside square
brackets describes the actions carried out by the PSA.
\begin{enumerate}
\item The PSA can move to various locations and measure environmental
factors. In the first exchange, reference resolution is used to
interpret the phrase ``all three decks'' as referring to the lower deck,
mid deck and flight deck. Plan optimization finds the most efficient
way to order the locations, and paraphrases the resulting plan back to
the user.
\begin{description}
\item[USER:]  Go to all three decks and measure carbon dioxide.
\item [PSA:] I will move to flight deck, mid deck and then lower deck
and I will measure carbon dioxide level, okay?
\item[USER:] Okay.
{\it [PSA moves to flight deck]}
\item [PSA:] The carbon dioxide level at the flight deck is one percent.
{\it[PSA moves to mid deck]}
\item[PSA:]
The carbon dioxide level at the mid deck is one percent.
{\it[PSA moves to lower deck]}
\item[PSA:]
The carbon dioxide level at the lower deck is one percent.
\end{description}

\item The next example involves resolution of ellipsis. The relevant
context is recovered from the preceding dialogue.
\begin{description}
\item[USER:]
Do the same for the pilot's seat.
\item[PSA:]
I will move to the pilot's seat and I will measure carbon dioxide level, okay?
\item[USER:]
Right.
{\it [PSA moves to pilot's seat]}
\item[PSA:]
The carbon dioxide level at the pilot's seat is one percent.
\end{description}

\item Resolution of ellipsis can involve filling in unspecified 
semantic parameters to a syntactically complete command. It is not
meaningful to talk about a sensor reading without reference to the
place at which that reading was carried out; in the following
example, resolution assumes that the location in question is the
one most recently mentioned in the dialogue. 

Note also that since the PSA does not have to move in order to carry
out the command, and hence can execute it immediately, the DM 
reasonably decides to skip the confirmation question. This 
decision is made based on the meta-output encoding the estimated cost
of executing the script.
\begin{description}
\item[USER:]
What was the carbon dioxide level at fifteen oh five according to the fixed
sensors?
\item[PSA:] According to the fixed sensors, at fifteen oh five the
carbon dioxide level at the pilot's seat was one percent.
\end{description}

\item The system encounters a user misconception, which results in a
``presupposition failure'' meta-output being produced during reference
resolution. It consequently decides to alert the user to this, rather
than attempt to carry out the command.
\begin{description}
\item[USER:]
Close both doors.
\item[PSA:]
There are in fact three of them.
\end{description}

\item The user utterance includes a pronoun, ``it''. The system
resolves ``it'' to ``the crew hatch'', and issues an appropriate
paraphrase in order to confirm that this is what the user meant.
\begin{description}
\item[USER:]
Go to crew hatch and close it.
\item[PSA:]
I will go to crew hatch and I will close crew hatch, okay?
\item[USER:]
Yeah.
{\it[PSA moves to crew hatch and closes it]}
\end{description}

\item The user asks the system to close the door, but it is unclear
which door they are referring to. An ``ambiguous reference''
meta-output is produced by reference resolution, resulting in the DM
issuing a clarification question.
\begin{description}
\item[USER:]
Close the door.
\item[PSA:]
Which door do you mean?
\end{description}

\item The user response is successfully interpreted by reference
resolution as an instance of ellipsis. Trying to close the crew hatch
a second time however causes a presupposition failure meta-output to
be generated during plan evaluation. The system informs the user of
this and takes no action.
\begin{description}
\item[USER:]
The crew hatch.
\item[PSA:]
The crew hatch is already closed.
\end{description}

\item The user presents three locations in a suboptimal order. Plan
evaluation finds a more efficient ordering.
\begin{description}
\item[USER:] Move to storage lockers, commander's seat and flight deck
and measure temperature.
\item[PSA:]
I will move to flight deck, commander's seat and then storage lockers and I will measure temperature, okay?
\item[USER:]
Sure.\\
{\it [PSA moves to flight deck, commander's seat and storage lockers in that order and reports temperature at each location]}
\end{description}

\item The user asks the system to repeat the previous command. Since the PSA is now starting from a different position, its most efficient plan is no longer the same, and it informs the user of this.
\begin{description}
\item[USER:]
Do that again.
\item[PSA:]
I will move to storage lockers, commander's seat and then flight deck and I will measure temperature, okay?
\item[USER:]
Yes.
{\it [PSA moves to storage lockers]}
\item[PSA:]
The temperature at the storage lockers is 19.9 degrees Celsius.
{\it [PSA starts moving to commander's seat]}
\end{description}

\item The user interrupts the PSA in the middle of the plan, and asks
it to go back to its last stopping place. Since ``stop'' may involve a
potentially hazardous condition, the interface obeys immediately,
without attempting to confirm first.
\begin{description}
\item[USER:]
Stop.
{\it [PSA stops]}
\item[USER:]
Go back.
{\it [PSA returns to storage lockers]}
\end{description}
\end{enumerate}

\section{Summary}
\label{Section:Conclusions}

We have described a generic architecture for spoken language dialogue
interfaces to semi-autonomous agents, based on the standard notion of
translating to successively deeper levels of representation. The novel
aspects of our scheme center around two ideas: first, that the final
output representations are best conceptualized not as logical
expressions but rather as programs in a scripting language; second,
that steps in the translation process should produce not only a simple
output, but also meta-information describing how the output was
produced. We have presented examples suggesting how several apparently
diverse types of dialogue behavior can be captured simply within our
framework, and outlined a prototype implementation of the scheme.

\bibliographystyle{acl}

\bibliography{rialist}

\end{document}